# Principle Driven Parameterized Fiber Model based on GPT-PINN Neural Network

Yubin Zang,[1] Boyu Hua,[2] Zhenzhou Tang,[2] Zhipeng Lin,[2] Fangzheng Zhang,[2] Simin Li,[2] Zuxing Zhang[1,*] and Hongwei Chen[3,*]

[1]*College of Electronic and Optical Engineering, Nanjing University of Posts and Telecommunications, Nanjing, China, 210023.*
[2]*College of Electronic and Information Engineering, Nanjing University of Aeronautics and Astronautics, Nanjing, China, 211106.*
[3]*Department of Electronic Engineering, Tsinghua University and Beijing National Research Center for Information Science and Technology, Beijing, China, 100084.*
*\*zxzhang@njupt.edu.cn    \*chenhw@tsinghua.edu.cn*

**Abstract:** In cater the need of Beyond 5G communications, large numbers of data driven artificial intelligence based fiber models has been put forward as to utilize artificial intelligence's regression ability to predict pulse evolution in fiber transmission at a much faster speed compared with the traditional split step Fourier method. In order to increase the physical interpretabiliy, principle driven fiber models have been proposed which inserts the Nonlinear Schodinger Equation into their loss functions. However, regardless of either principle driven or data driven models, they need to be re-trained the whole model under different transmission conditions. Unfortunately, this situation can be unavoidable when conducting the fiber communication optimization work. If the scale of different transmission conditions is large, then the whole model needs to be retrained large numbers of time with relatively large scale of parameters which may consume higher time costs. Computing efficiency will be dragged down as well. In order to address this problem, we propose the principle driven parameterized fiber model in this manuscript. This model breaks down the predicted NLSE solution with respect to one set of transmission condition into the linear combination of several eigen solutions which were outputted by each pre-trained principle driven fiber model via the reduced basis method. Therefore, the model can greatly alleviate the heavy burden of re-training since only the linear combination coefficients need to be found when changing the transmission condition. Not only strong physical interpretability can the model posses, but also higher computing efficiency can be obtained. The model's performance was demonstrated by the pulses with different shapes under 1000 different transmission conditions over the maximum 100 km fiber. The model's computational complexity is 0.0113% of split step Fourier method and 1% of the previously proposed principle driven fiber model.

## 1. Introduction

The rapidly developing artificially intelligence(AI) has profoundly and broadly changed the way of scientific researching [1-4]. There is no exception for fields of fiber modeling. In traditional fiber optics, fiber transmission properties are accurately described by Nonlinear Schrödinger Equations (NLSE) [5]. In order to solve it numerically, Split Step Fourier Method (SSFM) was put forward [6-7]. By taking the thoughts of 'differentiation' and 'decoupling' like other partial differential equations (PDE) solver, it firstly divides the whole fiber into multiple basic computing segments. In each segment, effects like attenuation, dispersion and nonlinearity can be viewed as separately affecting the pulse transmission. The final results can be obtained by finishing each segment's computing step by step. Though this method has become the most frequently adopted method in conventional fiber models thanks to its validity and accuracy, both disadvantages in computing speed and complexity uniformity under different circumstances can not be neglected.

Many researchers have proposed several new fiber models with the help of AI in order to address the above drawbacks of conventional models. Among them are BiLSTM model [8], GAN model [9-10], Multi-head Attention model [11-12] and etc. Most of these new AI based fiber models are data driven models in which large scale of data are collected in advance to catalyze the neural network to bridge the connection between waveform before and after transmission. In this case, these data driven models can posses high computing speed but discard the whole models from their physical backgrounds. Therefore, it will be convenient to utilized data driven fiber models for fiber transmission signal predicting under the condition that both signals before fiber transmission and after fiber transmission are easy to be collected. Without these data, these data driven models can no longer be effectively trained.

In order to improve the model's physical interpreter and the reliability on the pre-collected signals after the transmission, principle driven fiber transmission models were proposed [13-14]. Different from viewing the fiber transmission task as a pure mathematical regression work, these models utilize NLSE and its related constraints as loss functions. These models will then be optimized under the guidance of NLSE-related loss functions so that outputs in each training epoch will be checked the consistency of NLSE and its constraints. As a result, these models can not only obtain high physical interpretability, but also can be effectively trained even without the pre-collected transmitted data.

When dealing with the fiber transmission prediction for determined condition, both data driven and principle driven models are convenience and performs higher computing efficiency. However, in the next generation communications, fiber transmission system optimization work must be done where different transmission conditions can not be avoided. The typical characteristics of different transmission conditions is that parameters like fiber attenuation, dispersion or nonlinearity may vary. Under this circumstance, either data driven or principle driven models must be wholly retrained since the transmitted data may differ for data driven models and the coefficients in NLSE may vary for principle driven models. When the scale of different transmission conditions are larger, longer time will be consumed for re-training the whole AI based models which can greatly decrease the computing efficiency.

In order to both shorten the time of retraining and maintain the physical interpretabiliy, we propose the novelty parameterized principle driven fiber transmission model under the intuition of the generative pre-trained physics informed neural network (GPT-PINN) [15] in this manuscript. Unlike the previously proposed principle driven fiber model which deals with NLSE containing the constant parameter, this model performs as a solver for NLSE containing varying parameters including attenuation, dispersion, nonlinearity for different transmission conditions. This model views the predicted NLSE solution with respect to one set of parameter values being the results of several linearly combined eigen solutions which are outputted by the previously proposed principle driven models [11]. Therefore, only the combination ratio needs to be found for different circumstances. In this way, this model can save large amounts of time for retraining and computing complexity.

The manuscript will be developed into four main parts. All backgrounds of the fiber transmission models and the motivations of proposing this model will be described in the introduction part. Then, both theoretic principles, components, structures and computing flows of the parameterized fiber transmission model will be illustrated in detail in the second part. In order to demonstrates both model's predicting performances and gain in computational complexity, different tasks including single pulse input and multiple pluses input with different shapes will be illustrated. All results including the model's training behavior, predicting accuracy, application generalization and complexity comparisons will be shown and analyzed. Conclusions and further discussions will be developed in the last part.

## 2. Model Structures and Configurations

*2.1 Fiber Propagation Characteristics*

The propagation mechanism of optical fiber links is subject to the electromagnetic properties of light waves and the physical properties of optical fibers. When a waveform propagates in an optical fiber, it will be affected by factors such as loss, dispersion and nonlinearity in the optical fiber which can cause broadening and distortion.

Loss is one of the most common factors that can affect the evolution of the waveform inputs. Waveform's energy or power will decrease when propagating alongside the fiber. Dispersion, strictly speaking, chromatic dispersion, is the another factor affecting the evolution of the transmitted waveforms. The essence of chromatic dispersion is that different frequency components of one waveform will propagate at different group velocities through fiber which will finally cause waveform distortion at the end of fiber. The nonlinear effect is another important factors that can affect the waveform's shape. This effect is closely related with waveform's power, fiber's nonlinearity. Together with dispersion, this effect will make the waveform distort in a sophisticated way.

Though multiple effects can cause profound influences on sophisticated waveform evolution in the fiber, this process obey the Nonlinear Schrödinger Equation (NLSE) according to the theory of fiber optics.

$$iA_1\frac{\partial \psi}{\partial \zeta}+i\kappa_1 A_2\psi+\kappa_1 A_3\frac{\partial^2 \psi}{\kappa_2^2 \partial t^2}+i\kappa_1 A_4\frac{\partial^3 \psi}{\kappa_2^3 \partial t^3}+\kappa_1 A_5|\psi|^2\psi=0 \tag{1}$$

$$A_1=1 \quad A_2=\frac{\alpha L_D}{2} \quad A_3=-\frac{sign(\beta_2)}{2} \quad A_4=-\frac{\beta_3 L_D}{6T_0^3} \quad A_5=\frac{L_D}{L_{NL}} \tag{2}$$

$$z=L_{max}\zeta=\frac{L_{max}}{L_D}L_D\zeta=\kappa_1 L_D\zeta$$

$$\gamma=\frac{n_2\omega_c}{cA_{eff}}, \ T=T_{max}t=\frac{T_{max}}{T_0}T_0 t=\kappa_2 T_0\zeta \tag{3}$$

$$\Psi=\sqrt{P_0}\psi, \ L_D=-\frac{T_0^2}{\beta_2}, \ L_{NL}=\frac{1}{\gamma P_0}$$

where $\Psi=\Psi(T,z)$ is the solution of Eq. (1) whose diagram is a complex surface while $\psi=\psi(t,\zeta)$ represents the normalized solution. $T_0$ represents the full width of which the corresponding value equals 1/e of the pulse peak. $\alpha$ represents power attenuation per distance. $\beta_2$ and $\beta_3$ is the second and third order propagation constant which reflects dispersion and high order dispersion effects in the fiber respectively. $\gamma$ represents nonlinear coefficient. $n_2$, $\omega_c$, $c$ and $A_{eff}$ denotes nonlinear refractive index, central angular frequency of light, speed of light and effective cross-sectional area respectively.

## 2.2 Split Step Fourier Method

Since the NLSE is a partial differential equation, there is no analytical solution in most cases. In the actual modeling process, researchers often use the split-step Fourier method (SSFM) to numerically solve Eq. (1). SSFM belongs to forward iterative finite element methods. The idea of the method is to decompose a long transmission optical fiber into several computational finite elements called computing segment. In each segment, dispersion and non-linearity can be considered uncoupled so that these two effects act on the transmission signal in the optical fiber respectively. Under this circumstance, Eq. (1) can be written in operator form: Dispersion of the first half segment is firstly calculated. Waveform is transformed from time domain to frequency domain via Fast Fourier Transform (FFT). Then, quadrature phase shift term is multiplicated, and the whole waveform will be transformed from frequency domain back into the time domain via Inverse Fourier Transform (IFFT). The calculation of the nonlinear effect of the segment is calculated as the second part. Finally, dispersion of the second half of the segment is calculated which adopts the same procedures like the first half. This process is repeated segment by segment. Final results can be obtained when all segments are calculated.

It can be concluded from the above analysis that though SSFM has been most commonly used as the numerical solver of NLSE, its calculation accuracy strongly relates with the length of the segment. If the segment's length is determined too large, then the decouple approximation between dispersion and nonlinearity will cause large error in final results. If the segment's length is determined too small, then longer time will be consumed for calculation. Besides, it can also be seen that the complexity of the algorithm changes with the parameters of the specific optical fiber channel. When facing long-distance, high-power, and special optical fiber channels, the algorithm shows higher computing complexity and lower computational efficiency.

### 2.3 Principle Driven Fiber Model

In order to overcome the drawbacks of SSFM without losing the physical interpretabiliy like most data-driven AI based fiber models, the principle driven fiber model based on the physics informed neural network (PINN) have been put forward. It views AI as a powerful equation solver for NLSE which is essentially different from other data driven fiber models.

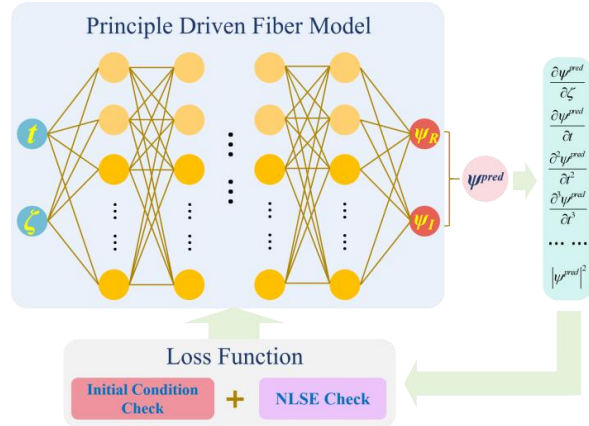

Fig. 1. The structure of principle driven fiber model

The universal structure of the principle driven fiber model is a multi-layer neural network containing one input layer, several hidden layers and one output layer as is shown in Fig. 1. As can be observed from Eq. (1)-(3), both variables ($t,\zeta$) span a plane which contentiously takes value from $R^2$ space. However, the principle driven fiber model can only execute the computing point by point. Therefore, the continuous two-dimensional plane spanned by ($t,\zeta$) needs to be discretized before entering into the principle driven fiber model. Under this circumstance, there are two input neurons in the model with one takes the discrete $t$ value and the other takes the discrete $\zeta$ value. Hidden layers are responsible for mapping the inputs into the corresponding solution value $\psi$. Since the solution value is a complex number, there are two neurons in the output layer of the fiber model with one outputting the real part and the other outputting the imaginary part. The computations between each layer is linear multiplications and $Tanh()$ activations.

Unlike data driven models, loss function is crucial and should be carefully designed in the principle fiber model since it directly relates with the training converging performance. The basic idea of loss function determination is to check the distance between the model's prediction and the NLSE solution by substituting the model prediction into the NLSE and calculate the residual error. In this case, NLSE residual error becomes one part of the loss function. Apart from this, the input waveform, as the initial condition, also affects the fiber transmission. Therefore, the normalized mean square error between model's prediction when $\zeta$ equals zero and the original input waveform should be formed as the second part of the loss function. In this way, the loss function of the principle driven fiber model can be determined as

$$L^F = \left\{ \begin{array}{l} \dfrac{1}{N_E}\sum_{k=1}^{N_E}\left\| A_1\dfrac{\partial \psi_k^{pred}}{\partial \zeta} + i\kappa_1 A_2 \psi_k^{pred} + \kappa_1 A_3 \dfrac{\partial^2 \psi_k^{pred}}{\kappa_2^2 \partial t^2} + i\kappa_1 A_4 \dfrac{\partial^3 \psi_k^{pred}}{\kappa_2^3 \partial t^3} + \kappa_1 A_5 \left|\psi_k^{pred}\right|^2 \psi_k^{pred} \right\|^2 \\ + \dfrac{1}{N_{ini}}\sum_{k=1}^{N_{ini}}\left\| \psi_k^{pred}(\zeta=0,t) - \psi_k(\zeta=0,t) \right\|^2 \end{array} \right\} \quad (4)$$

where $L^F$ represents the loss function of the principle fiber model. $\psi^{pred}$ represents the prediction of the model and it consists of the real part and imaginary part which can be expressed as $\psi^{pred}=\psi_R^{pred}+i\psi_I^{pred}$. In this way, Eq. (4) can be further described as

$$L^F = \left\{ \begin{array}{l} \dfrac{1}{N_E}\sum_{k=1}^{N_E}\left\| \begin{array}{l} -A_1\dfrac{\partial \psi_{Ik}^{pred}}{\partial \zeta} - \kappa_1 A_2 \psi_{Ik}^{pred} + \kappa_1 A_3 \dfrac{\partial^2 \psi_{Rk}^{pred}}{\kappa_2^2 \partial t^2} \\ -\kappa_1 A_4 \dfrac{\partial^3 \psi_{Ik}^{pred}}{\kappa_2^3 \partial t^3} + \kappa_1 A_5 \left[\left(\psi_{Rk}^{pred}\right)^2 + \left(\psi_{Ik}^{pred}\right)^2\right]\psi_{Rk}^{pred} \end{array} \right\|^2 \\ + \dfrac{1}{N_E}\sum_{k=1}^{N_E}\left\| \begin{array}{l} A_1\dfrac{\partial \psi_{Rk}^{pred}}{\partial \zeta} + \kappa_1 A_2 \psi_{Rk}^{pred} + \kappa_1 A_3 \dfrac{\partial^2 \psi_{Ik}^{pred}}{\kappa_2^2 \partial t^2} \\ +\kappa_1 A_4 \dfrac{\partial^3 \psi_{Rk}^{pred}}{\kappa_2^3 \partial t^3} + \kappa_1 A_5 \left[\left(\psi_{Rk}^{pred}\right)^2 + \left(\psi_{Ik}^{pred}\right)^2\right]\psi_{Ik}^{pred} \end{array} \right\|^2 \\ + \dfrac{1}{N_{ini}}\sum_{k=1}^{N_{ini}}\left\| \psi_{Rk}^{pred}(\zeta=0,t) - \psi_{Rk}(\zeta=0,t) \right\|^2 \\ + \dfrac{1}{N_{ini}}\sum_{k=1}^{N_{ini}}\left\| \psi_{Ik}^{pred}(\zeta=0,t) - \psi_{Ik}(\zeta=0,t) \right\|^2 \end{array} \right\} \quad (5)$$

The hyper-parameters in the principle driven fiber model includes the number of hidden layers, the scale of neurons in each layer etc which should be determined in advance based on the extent of complexity of the task.

In all, the principle fiber model proposed before indeed can posses great prediction accuracy without losing too much physical background. However, it is still inconvenient when parameters such as either attenuation, dispersion or non-linearity changes because the whole model needs to be re-trained and can cost large amount of time. Therefore, actions should be taken to develop more universal solutions of the parameterized NLSE on the basis of the prediction of the proposed principle driven fiber model.

*2.4 Reduced Basis Expansion Method*

Reduced basis expansion method is probably one of the most feasible methods for further developing the more universal solutions based on several pre-predicted solutions. The thinking behind the reduced basis expansion method stems from the knowledge of basis expansion. Looking back towards either linear algebra or signal processing theories, basis expansion always performs an important role in decomposing one sophisticated generalized object into several simple but typical selected components. As can be seen from Fig. 2, vectors can be decomposed into coordinate basis in analytic geometry; periodic signals can be decomposed into several sine or cosine functions in Fourier series; optical field distribution at the cross section of fiber can be decomposed into multiple modes in the theory of electromagnetic field. The common idea behind is to utilize the linear combinations of basis, either finite or infinite, to express the universal or general object. Similarly, the reduced basis expansion method views the universal solutions of parameterized NLSE as the linear combinations of several typically selected special solutions called 'eigen solutions'. Under this circumstance, both linear combination coefficients and the typically selected eigen solutions need to be found in order to obtain the universal solutions of parameterized NLSE.

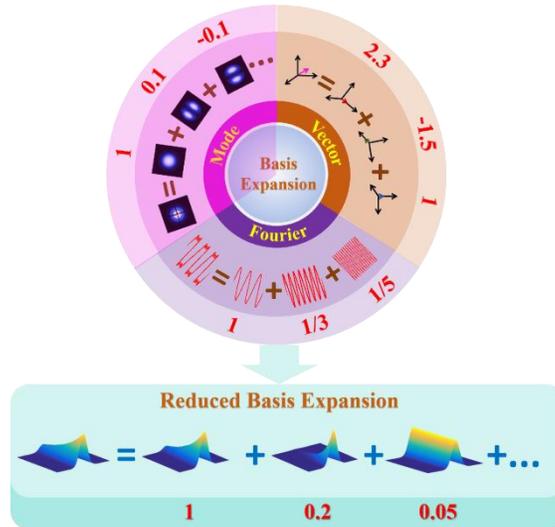

Fig. 2. The structure of principle driven fiber model

Unlike the conventional basis expansion method as the examples illustrated above that all basis shall meet the criterion of orthogonality and completeness, the eigen solutions from the reduced basis expansion method are always non-orthogonal and finite scaled which means that no universal solutions can be perfectly and accurately forms by using eigen solutions without any approximation errors. Nonetheless, the reduced basis expansion method is still effective as long as the errors of the linearly expressed universal solutions are acceptable. In another word, if both eigen solutions and linear combination coefficients can be appropriately found, then this method would be a relatively good approximation of the theoretic universal solutions of parameterized NLSE.

Greedy algorithm is used to find both eigen solutions and linear combination coefficients in the reduced basis expansion method. At the start of greedy algorithm, one set of values in the parameter space of the parameterized NLSE is pre-determined. Then, the principle driven fiber model is trained to find the predicted solution of NLSE with this set of parameters. The predicted solution of this well trained principle driven model forms as the first eigen solution. Then, this eigen solution is utilized to approximate all solutions stems from the parameterized NLSE in the parameter space. The criterion for approximation performance by using the first eigen solution is to substitute the approximation results into both NLSE and initial condition to calculate the errors. Afterwards, the set of parameters whose error is the largest is utilized to train the principle driven fiber model to find the second eigen solution. The second eigen solution will together with the first eigen solution to find the best coefficients which can result in the minimum errors for all solutions with respect to the parameter space of the parameterized NLSE. The process repeats unless the acceptable approximation of the universal solutions are found.

Two important information should be taken extra attention. First, gradient descend method is utilized to find the combination coefficients after each eigen solution is found. Therefore, both values and gradients of the loss function of approximation results with respect to the linear combination coefficients should be mathematically defined and calculated. Second, if one set of parameter is chosen to be the resource to find the eigen solution, then this set of parameter will no longer exist in the parameter space for testing the approximation performance since the eigen solution alone is the best approximation for itself.

Under the procedures illustrated above, for the *p*-th iteration, the approximated universal solutions of the parameterized NLSE can be expressed as

$$\Phi_p^{pred} = \sum_{m=1}^{p} c_m \psi_m^{pred} = \sum_{m=1}^{p} c_m \psi_{Rm}^{pred} + i \sum_{m=1}^{p} c_m \psi_{Im}^{pred} \tag{6}$$

in which *c* represents the linear combination coefficients and take real values. By substitution Eq. (6) into Eq. (4), the loss function for the parameterized NLSE can be found.

$$L^{PF} = \left\{ \begin{array}{l} \dfrac{1}{N_E} \sum_{k=1}^{N_E} \left\| \begin{array}{l} -A_1 \dfrac{\partial}{\partial \zeta} \sum_{m=1}^{p} c_m \psi_{Im,k}^{pred} - \kappa_1 A_2 \sum_{m=1}^{p} c_m \psi_{Im,k}^{pred} \\ + \kappa_1 A_3 \dfrac{\partial^2}{\kappa_2^2 \partial t^2} \sum_{m=1}^{p} c_m \psi_{Rm,k}^{pred} - \kappa_1 A_4 \dfrac{\partial^3}{\kappa_2^3 \partial t^3} \sum_{m=1}^{p} c_m \psi_{Im,k}^{pred} \\ + \kappa_1 A_5 \left[ \left( \sum_{m=1}^{p} c_m \psi_{Rm,k}^{pred} \right)^2 + \left( \sum_{m=1}^{p} c_m \psi_{Im,k}^{pred} \right)^2 \right] \sum_{m=1}^{p} c_m \psi_{Rm,k}^{pred} \end{array} \right\|^2 \\[2ex] + \dfrac{1}{N_E} \sum_{k=1}^{N_E} \left\| \begin{array}{l} A_1 \dfrac{\partial}{\partial \zeta} \sum_{m=1}^{p} c_m \psi_{Rm,k}^{pred} + \kappa_1 A_2 \sum_{m=1}^{p} c_m \psi_{Rm,k}^{pred} \\ + \kappa_1 A_3 \dfrac{\partial^2}{\kappa_2^2 \partial t^2} \sum_{m=1}^{p} c_m \psi_{Im,k}^{pred} + \kappa_1 A_4 \dfrac{\partial^3}{\kappa_2^3 \partial t^3} \sum_{m=1}^{p} c_m \psi_{Rm,k}^{pred} \\ + \kappa_1 A_5 \left[ \left( \sum_{m=1}^{p} c_m \psi_{Rm,k}^{pred} \right)^2 + \left( \sum_{m=1}^{p} c_m \psi_{Im,k}^{pred} \right)^2 \right] \sum_{m=1}^{p} c_m \psi_{Im,k}^{pred} \end{array} \right\|^2 \\[2ex] + \dfrac{1}{N_{ini}} \sum_{k=1}^{N_{ini}} \left\| \sum_{m=1}^{p} c_m \psi_{Rm,k}^{pred}(\zeta=0,t) - \psi_{Rk}(\zeta=0,t) \right\|^2 \\[2ex] + \dfrac{1}{N_{ini}} \sum_{k=1}^{N_{ini}} \left\| \sum_{m=1}^{p} c_m \psi_{Im,k}^{pred}(\zeta=0,t) - \psi_{Ik}(\zeta=0,t) \right\|^2 \end{array} \right\} \tag{7}$$

By taking the derivatives of $L^{PF}$ with respect to the m-th linear combination coefficient, one can easily find the gradient which is shown in Eq. (9).

$$\left( \nabla L^{PF} \right)_{c_m}^{p} = \left\{ \begin{array}{l} \dfrac{2}{N_E} \sum_{k=1}^{N_E} \left[ \begin{array}{l} \left( \begin{array}{l} -A_1 \dfrac{\partial}{\partial \zeta} \sum_{m=1}^{p} c_m \psi_{Imk}^{pred} - \kappa_1 A_2 \sum_{m=1}^{p} c_m \psi_{Imk}^{pred} \\ + \kappa_1 A_3 \dfrac{\partial^2}{\kappa_2^2 \partial t^2} \sum_{m=1}^{p} c_m \psi_{Rmk}^{pred} - \kappa_1 A_4 \dfrac{\partial^3}{\kappa_2^3 \partial t^3} \sum_{m=1}^{p} c_m \psi_{Imk}^{pred} \\ + \kappa_1 A_5 \left[ \left( \sum_{m=1}^{p} c_m \psi_{Rmk}^{pred} \right)^2 + \left( \sum_{m=1}^{p} c_m \psi_{Imk}^{pred} \right)^2 \right] \sum_{m=1}^{p} c_m \psi_{Rmk}^{pred} \end{array} \right) \\ \times \left( \begin{array}{l} -A_1 \dfrac{\partial \psi_{Imk}^{pred}}{\partial \zeta} - \kappa_1 A_2 \psi_{Imk}^{pred} + \kappa_1 A_3 \dfrac{\partial^2 \psi_{Rmk}^{pred}}{\kappa_2^2 \partial t^2} - \kappa_1 A_4 \dfrac{\partial^3 \psi_{Imk}^{pred}}{\kappa_2^3 \partial t^3} \\ + \kappa_1 A_5 \left[ \left( \sum_{m=1}^{p} c_m \psi_{Rmk}^{pred} \right)^2 + \left( \sum_{m=1}^{p} c_m \psi_{Imk}^{pred} \right)^2 \right] \psi_{Rmk}^{pred} \\ + 2\kappa_1 A_5 \sum_{m=1}^{p} c_m \psi_{Rmk}^{pred} \left( \psi_{Rmk}^{pred} + \psi_{Imk}^{pred} \right) \end{array} \right) \end{array} \right] \\[2ex] + \dfrac{1}{N_E} \sum_{k=1}^{N_E} \left[ \begin{array}{l} \left( \begin{array}{l} A_1 \dfrac{\partial}{\partial \zeta} \sum_{m=1}^{p} c_m \psi_{Rmk}^{pred} + \kappa_1 A_2 \sum_{m=1}^{p} c_m \psi_{Rmk}^{pred} \\ + \kappa_1 A_3 \dfrac{\partial^2}{\kappa_2^2 \partial t^2} \sum_{m=1}^{p} c_m \psi_{Imk}^{pred} + \kappa_1 A_4 \dfrac{\partial^3}{\kappa_2^3 \partial t^3} \sum_{m=1}^{p} c_m \psi_{Rmk}^{pred} \\ + \kappa_1 A_5 \left[ \left( \sum_{m=1}^{p} c_m \psi_{Rmk}^{pred} \right)^2 + \left( \sum_{m=1}^{p} c_m \psi_{Imk}^{pred} \right)^2 \right] \sum_{m=1}^{p} c_m \psi_{Imk}^{pred} \end{array} \right) \\ \times \left( \begin{array}{l} A_1 \dfrac{\partial \psi_{Rmk}^{pred}}{\partial \zeta} + \kappa_1 A_2 \psi_{Rmk}^{pred} + \kappa_1 A_3 \dfrac{\partial^2 \psi_{Imk}^{pred}}{\kappa_2^2 \partial t^2} + \kappa_1 A_4 \dfrac{\partial^3 \psi_{Rmk}^{pred}}{\kappa_2^3 \partial t^3} \\ + \kappa_1 A_5 \left[ \left( \sum_{m=1}^{p} c_m \psi_{Rmk}^{pred} \right)^2 + \left( \sum_{m=1}^{p} c_m \psi_{Imk}^{pred} \right)^2 \right] \psi_{Imk}^{pred} \\ + 2\kappa_1 A_5 \sum_{m=1}^{p} c_m \psi_{Imk}^{pred} \left( \psi_{Rmk}^{pred} + \psi_{Imk}^{pred} \right) \end{array} \right) \end{array} \right] \\[2ex] + \dfrac{2}{N_{ini}} \sum_{k=1}^{N_{ini}} \left( \sum_{m=1}^{p} c_m \psi_{Rmk}^{pred}(\zeta=0,t) - \psi_{Rk}(\zeta=0,t) \right) \psi_{Rmk}^{pred}(\zeta=0,t) + \dfrac{2}{N_{ini}} \sum_{k=1}^{N_{ini}} \left( \sum_{m=1}^{p} c_m \psi_{Imk}^{pred}(\zeta=0,t) - \psi_{Ik}(\zeta=0,t) \right) \psi_{Imk}^{pred}(\zeta=0,t) \end{array} \right\} \tag{8}$$

Then the linear combination coefficient can be updated by utilizing Eq. (8) as

$$c_m^t = c_m^t - lr \times \left( \nabla L^{PF} \right)_{c_m}^{t} \tag{9}$$

*2.5 Parameterized Principle Driven Fiber Models*

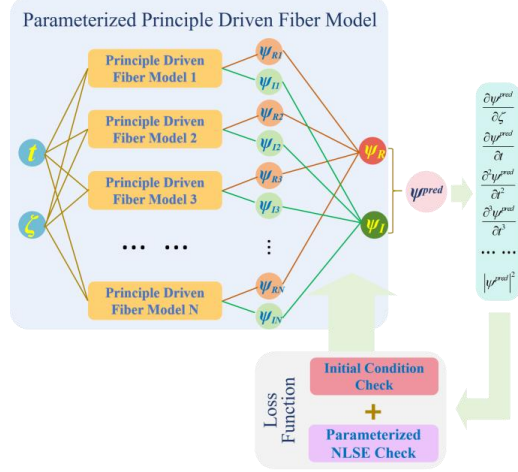

Fig. 3 The structure of the parameterized principle driven fiber model

As is illustrated above, the parameterized principle driven fiber model consists of both the principle driven fiber model and the reduced basis expansion method. For the training of this parameterized model, the reduced basis expansion method plays the role of finding the eigen solutions and the linear combination coefficients. For each eigen solution except the first, those set of parameters whose errors are the largest is utilized to train the principle fiber model. Then, the predicted solution will become the new eigen solution together with previous eigen solutions to train the best linear combination coefficients in order to minimize the overall loss towards the whole parameter space shown in Eq. (10) so as to reach the best approximation of the universal solution of the parameterized NLSE.

Without losing the generality, pulses with Gaussian, Sech and SuperGaussian shape are utilized to demonstrate the parameterized principle driven fiber model. For the disecretized of the independent variables $(t, \zeta)$, $t$ ranges from -1 to 1 with 100 points while $\zeta$ ranges from 0 to 1 with 101 points. The initial condition is also discretized into 100 points. For the parameterized space of the fiber, both attenuation, dispersion, either second-order or third-order and non-linearity can be ranged freely. All ranges and discretized points can be found in Table 1.

**Table 1. PARAMETER CONFIGURATIONS**

| PARAMETERS | RANGES | POINTS |
| --- | --- | --- |
| $t$ | [-1, 1] | 100 |
| $\zeta$ | [0, 1] | 101 |
| INITIAL CONDITIONS | GAUSSIAN, SECH, SUPERGAUSSIAN | 100 |
| $\alpha$ | $[0, 4.605 \times 10^{-5}]$/m | 10 |
| $\beta_2$ | $[-2 \times 10^{-26}, 2 \times 10^{-26}]$ s$^2$/m | 10 |
| $\beta_3$ | $-2 \times 10^{-38}$ m$^{-3}$ s$^3$/m | 1 |
| $n_2$ | $[-2.6 \times 10^{-22}, 2.6 \times 10^{-20}]$ m$^2$/W | 10 |
| $A_{eff}$ | $8 \times 10^{-11}$ m$^2$ | 1 |
| $\lambda$ | 1.55 μm | 1 |

When it comes to the configuration of the principle driven fiber model which can be seen in Table 2, for both single and multiple pulse predicting, the scale of hidden layers is determined to be 4 layers, each with 100 neurons. Early stopping and maximum training epochs determination is of equal significance since it can affect the final accuracy of the results. The early stopping error and maximum training epoch equals to be $10^{-5}$ and 60000 respectively. The .number of eigen solutions is determined to be 10. As for the configurations of the reduced basis expansion method, the maximum number of basis taken for estimating the universal solution is decided to be 10.

**Table 2. PARAMETERIZED FIBER MODEL CONFIGURATIONS**

| CONFIGURATION | | VALUE |
|---|---|---|
| EACH PRINCIPLE FIBER MODEL | TYPE | FULLY CONNECTED |
| | LAYER | [2,100,100,100,100,2] |
| | STOP CRITERION | Loss lower than $10^{-5}$ |
| | | Epochs larger than 60000 |
| REDUCED BASIS EXPANSION METHOD | NUMBER. OF EIGEN SOLUTIONS | 10 |

**Table 3. INPUT CONDITIONS AND SOLUTIONS CONFIGURATIONS**

| CONFIGURATION | | | VALUE |
|---|---|---|---|
| CENTRAL WAVELENGTH | | | 1550 nm |
| MAXIMUM POWER $P_{max}$ | | | 1 mW |
| SINGLE PULSE INPUT | GAUSSIAN | 1/e full width $T_0$ | $1/\sqrt{10}$ ns |
| | | Time range | $[-5/\sqrt{10}, 5/\sqrt{10}]$ ns |
| | | Distance range | [-100, 100] km |
| | SECH | 1/e full width $T_0$ | $1/\sqrt{10}$ ns |
| | | Time range | $[-5/\sqrt{10}, 5/\sqrt{10}]$ ns |
| | | Distance range | [-100, 100] km |
| | SUPER-GAUSSIAN | Order | 4 |
| | | 1/e full width $T_0$ | $1/\sqrt{10}$ ns |
| | | Time range | $[-5/\sqrt{10}, 5/\sqrt{10}]$ ns |
| | | Distance range | [-100, 100] km |
| MULTI PULSES INPUT | NUMBER. OF PULSES | | 4 |
| | PEAKS LOCATIONS | | $-10.5/\sqrt{10}$ ns |
| | GAUSSIAN | 1/e full width $T_0$ | $1/\sqrt{10}$ ns |
| | | Time range | $[-15/\sqrt{10}, 15/\sqrt{10}]$ ns |
| | | Distance range | [-100, 100] km |
| | SECH | 1/e full width $T_0$ | $1/\sqrt{10}$ ns |
| | | Time range | $[-15/\sqrt{10}, 15/\sqrt{10}]$ ns |
| | | Distance range | [-100, 100] km |
| | SUPER-GAUSSIAN | Order | 4 |
| | | 1/e full width $T_0$ | $1/\sqrt{10}$ ns |
| | | Time range | $[-15/\sqrt{10}, 15/\sqrt{10}]$ ns |
| | | Distance range | [-100, 100] km |

## 3. Simulation results and analysis

### 3.1 Training of the Principle Driven Fiber Model

The training of the principle driven fiber model aims at obtaining eigen solutions for the parameterized NLSE. Apart from the first set of parameters which is specified in advance, other sets of parameters come from the cases whose error is the largest error. For the training of the principle driven model, adaptive stochastic gradient descend method (ADAM) [16] is utilized as to better training the model. The adaptive learning rate controlled by the decaying coefficient in the method can let the model quickly find the local minimum area at first and prematurely search for the local minima at the final few epochs

**Table 4. EIGEN SOLUTIONS FOR GAUSSIAN PULSE INPUT**

| | CONVERGENCE | EIGEN SOLUTIONS | | CONVERGENCE | EIGEN SOLUTIONS |
|---|---|---|---|---|---|
| 1 | 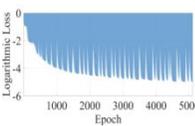 | 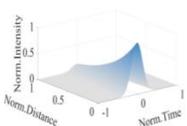 | 6 | 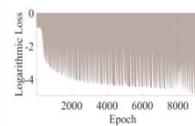 | 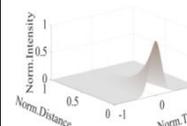 |
| 2 | 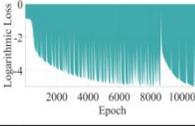 | 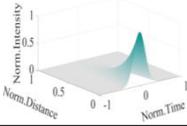 | 7 | 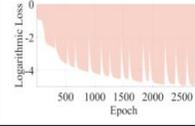 | 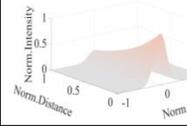 |
| 3 | 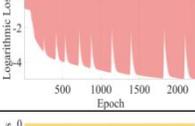 | 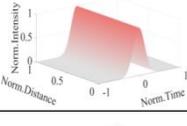 | 8 | 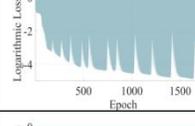 | 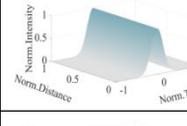 |
| 4 | 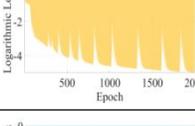 | 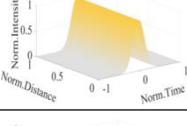 | 9 | 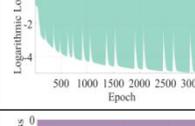 | 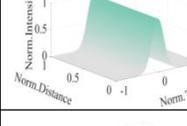 |
| 5 | 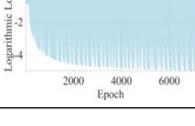 | 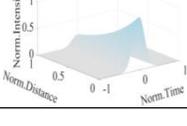 | 10 | 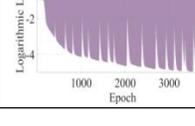 | 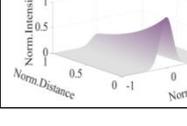 |

Since there are multiple models for training, only the convergence performance of training the first principle driven fiber model is given in Table 4. As can be observed, the errors both containing the NLSE residue and the initial condition difference descends quickly in the first 1000 epochs. The pace of descending slows down as the epoch goes which in turn demonstrate the property of ADAM optimizer. The training will end by either the model meets the error criterion or the maximum training epoch pre-determined before training. In most cases, $10^{-4}$ can be a relatively accurate error.

It can also be observed from the convergence diagram that multiple fluctuations exist which cause the whole curve being unsmooth. This phenomenon is proper and it indicates that it can be relatively challenging for the model to map both time and distance coordinate into waveform transmitted.

When compared with the convergence performances of finding each eigen solution as is shown in the left column in Table 4, both convergence speed and final errors are different.

This indicates that different sets of parameters in the parameterized NLSE has different difficulties in training the model to converge.

### 3.2 Training of the Reduced Basis Expansion

After the principle driven model is appropriately trained, the solution of this set of parameters will form new eigen solution. The linear combination coefficients will be trained as to find the best linear combinations of all eigen solutions obtained to in the current stage approximate the solutions with respect to other sets of parameters.

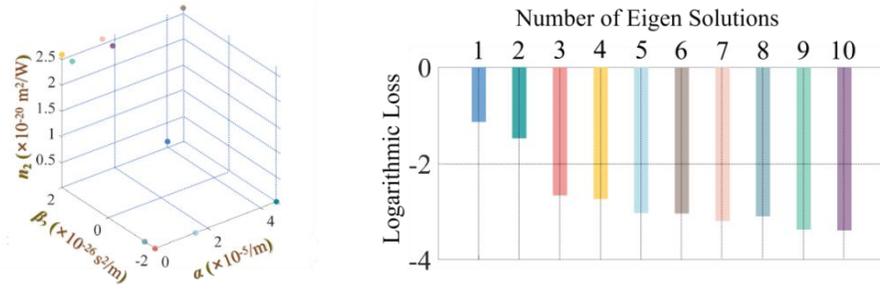

(a) Eigen solutions' parameters　　　　(b) Loss decreases as the number of eigen solution increase
Fig. 4 Results of the reduced basis expansion method for single Gaussian pulse as input

Fig. 4(a) shows ten sets of parameters utilized to find the corresponding eigen solutions while Fig. 4(b) shows the model's convergence situation with respect to the different numbers of eigen solutions for the single Gaussian pulse as input. 10 different colors in both subfigures mark the corresponding order of the 10 eigen solutions. In total two important conclusions can be drawn. First, the prediction loss indeed decreases as the number of eigen solutions increases. This is in consistent with the trend that more meticulous approximation can be reached when the number of basis increases according to the reduced basis expansion method. Second, the pace of convergence tends to slow down when adding more eigen solutions. This phenomenon can also be explained when conducting analogy work towards Fourier series that the latter-added eigen solutions perform as higher-order components which make less contributions than those first few eigen solutions. Besides, other interesting fact can also be found by carefully observing the upper Fig. 4(a) that most of the parameters selected for training of the eigen solutions are located at the boundary of the whole parameter space such as the $3^{rd}$ and $9^{th}$ set of parameters. This is because boundary of the parameter space represents relatively extreme transmission conditions which can cause larger difficulty for the principle driven fiber model to converge.

### 3.3 Performances of the Parameterized Fiber Model

The final parameterized principle driven fiber model utilizes the predicted solutions from the previously principle driven fiber models to form the final universal solutions with respect to the whole parameter space shown and depicted in both Table 1 and Fig. 4. Since single pulse with Gaussian, Sech and SuperGaussian are all utilized for demonstration, Fig. 5 shows the model's outputs for randomly selected 6 sets of parameters in the parameter space for the input of the single pulse with all three shapes. It can be concluded though approximation error exists, the reduced basis expansion method can still find the best linear coefficients to well organize the eigen solutions to form the final solutions for different transmission parameters so that the prediction accuracy can still remain high.

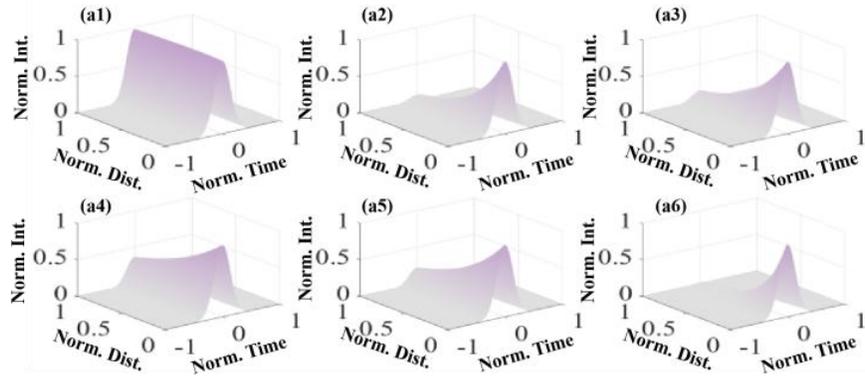

(a) Gaussian pulse

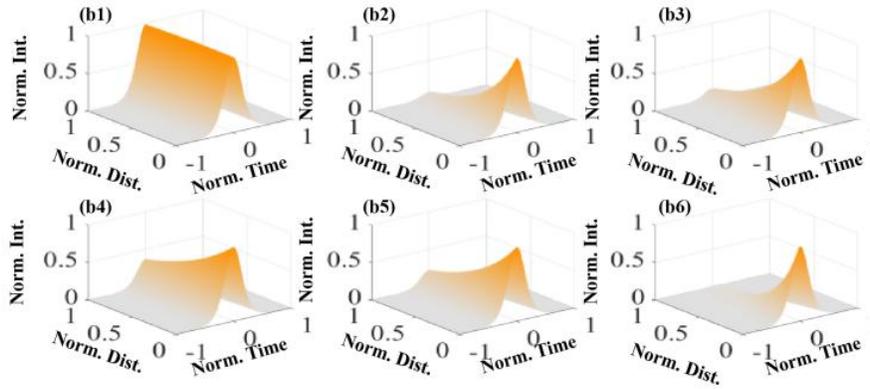

(b). Sech pulse

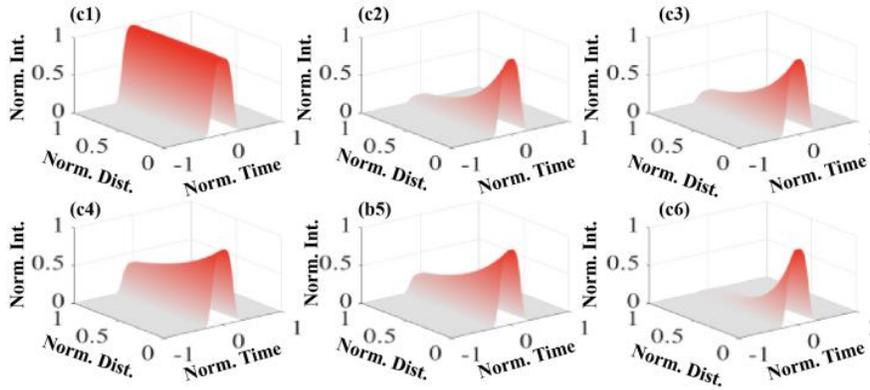

(c). SuperGaussian pulse

Fig. 5. Single pulse input
Parameter configurations.
(1) $\alpha=0$/m, $\beta_2=6.67\times10^{-27}$s$^2$/m, $n_2=8.84\times10^{-21}$m$^2$/W
(2) $\alpha=2.5584\times10^{-5}$/m, $\beta_2=-1.11\times10^{-26}$s$^2$/m, $n_2=5.98\times10^{-21}$m$^2$/W
(3) $\alpha=2.0476\times10^{-5}$/m, $\beta_2=1.56\times10^{-26}$s$^2$/m, $n_2=8,84\times10^{-21}$m$^2$/W
(4) $\alpha=1.0234\times10^{-5}$/m, $\beta_2=-1.11\times10^{-26}$s$^2$/m, $n_2=2.60\times10^{-20}$m$^2$/W
(5) $\alpha=1.5350\times10^{-5}$/m, $\beta_2=-6.67\times10^{-27}$s$^2$/m, $n_2=8.84\times10^{-21}$m$^2$/W
(6) $\alpha=4.0934\times10^{-5}$/m, $\beta_2=2.00\times10^{-26}$s$^2$/m, $n_2=5.98\times10^{-21}$m$^2$/W

## 3.4 Parameterized Fiber Model's training and testing performances for multiple pulses input

Though detailed training and testing performances of the model are illustrated and shown in detail for single pulse input, the prediction of multiple pulses input are also conducted whose configurations can be found in Table 1 and Table 2.

Like single pulse input, convergence performances for finding 10 eigen solutions when multiple Gaussian pulses being inputted are shown in Table 5. When compared with Table 4, it can be obviously found that the overall average number of epochs that can meet the stop criterion is larger for multiple pulses input. It is because multiple pulses posses more sophisticated waveform evolutions than single pulse thus will increase the difficulty for the principle model to converge to the expected state.

**Table 5. EIGEN SOLUTIONS FOR MULTIPLE GAUSSIAN PULSES INPUT**

| | CONVERGENCE | EIGEN SOLUTIONS | | CONVERGENCE | EIGEN SOLUTIONS |
|---|---|---|---|---|---|
| 1 | | | 6 | | |
| 2 | | | 7 | | |
| 3 | | | 8 | | |
| 4 | | | 9 | | |
| 5 | | | 10 | | |

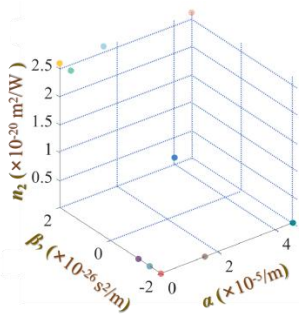
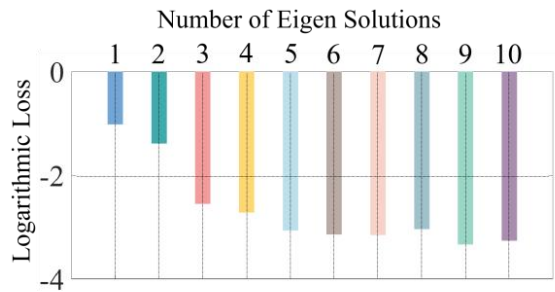

(a) Eigen solutions' parameters      (b) Loss decreases as the number of eigen solution increase

Fig. 6  Results of the reduced basis expansion method for multiple Gaussian pulses as input

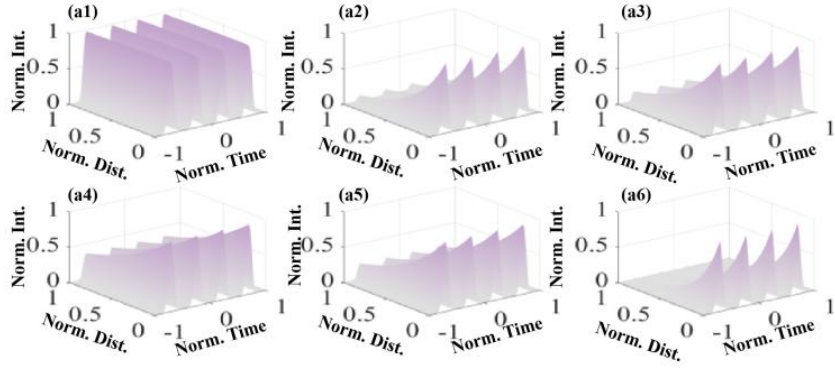

(a). Gaussian pulse.

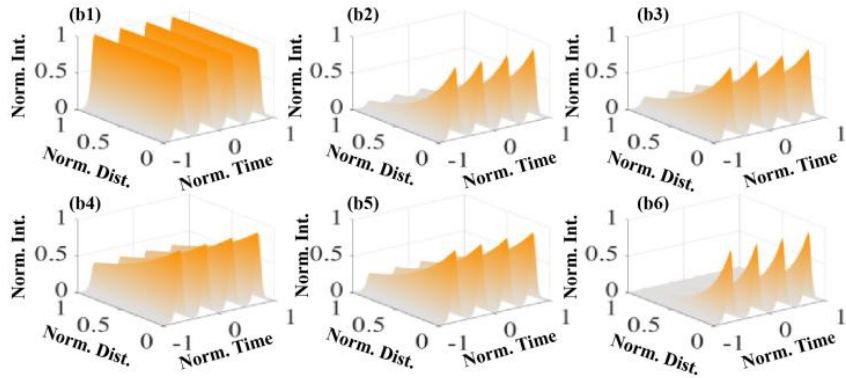

(b). Sech pulse

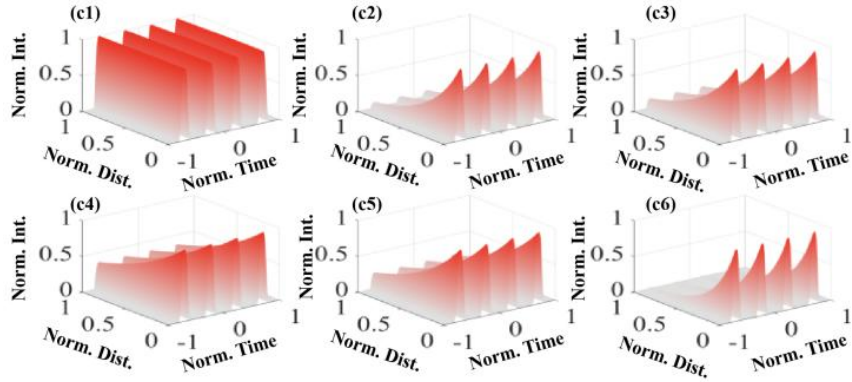

(c). SuperGaussian pulse
Fig. 7. Multiple pulses input.
Parameter configurations.
(1) $\alpha=0$/m, $\beta_2=6.67\times10^{-27}s^2$/m, $n_2=8.84\times10^{-21}$m2/W
(2) $\alpha=2.5584\times10^{-5}$/m, $\beta_2=-1.11\times10^{-26}s^2$/m, $n_2=5.98\times10^{-21}$m$^2$/W
(3) $\alpha=2.0476\times10^{-5}$/m, $\beta_2=1.56\times10^{-26}s^2$/m, $n_2=8,84\times10^{-21}$m$^2$/W
(4) $\alpha=1.0234\times10^{-5}$/m, $\beta_2=-1.11\times10^{-26}s^2$/m, $n_2=2.60\times10^{-20}$m$^2$/W
(5) $\alpha=1.5350\times10^{-5}$/m, $\beta_2=-6.67\times10^{-27}s^2$/m, $n_2=8.84\times10^{-21}$m$^2$/W
(6) $\alpha=4.0934\times10^{-5}$/m, $\beta2=2.00\times10^{-26}s^2$/m, $n_2=5.98\times10^{-21}$m$^2$/W

Like single pulse input, Fig. 7 depicts the prediction of parameterized principle driven model for multiple pulses input with the shape of Gaussian, Sech and SuperGaussian. Generally, both errors and model complexity are larger for multiple pulses input. It is possibly because multiple pulses present more sophisticated transmission behaviors. Given the same physical characteristics of fiber, more interference between each pulse in multiple pulses input must be taken into the account for the model. Therefore, the model for multiple pulses input tend to be designed more larger so as to better capture the extra transmission features the single pulse input does not behave.

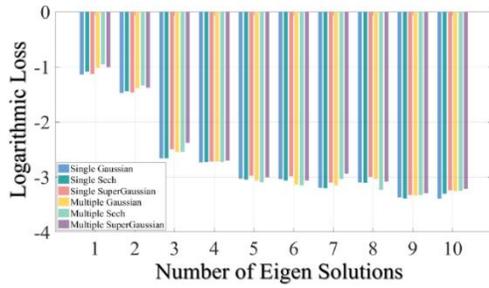 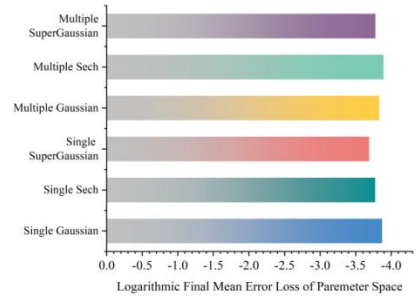

Fig. 8. Error analysis of the model          Fig. 9. Logarithmic final mean error

The total prediction loss for the model with regard to different input configurations is depicted in Fig. 8. When it comes to the prediction error analysis as is shown in Fig. 8, there exist several effective factors. First, different initial conditions can affect the approximation error. In general, rapid changed waveform may result in larger prediction error since it is hard for the same scale of the model to conduct regression for those signals whose bandwidth is wide. Second, the scale of the model can affect the prediction error. More sophisticated structured principle driven fiber models inside the parameterized fiber model can capture more transmission characteristics thus will lead to more accurate basis. More eigen solutions will also tend to result in less approximation error.

The final prediction performance is depicted in Fig. 9. Here, prediction error is measured by the value of loss function containing both NLSE residue loss and initial condition loss. Different colors which are in consistent with those in Fig. 8 mark different types of pulses input. For each type of pulse, average value of the the prediction over time and distance coordinates under 1000 different parameters is computed. Under the appropriate training, all prediction errors are less than $10^{-3.5}$ which are relatively high from the perspective of optical engineering.

More detailed error distributions over different parameters can be seen from the 3 dimensional scatter plot in Fig. 10. Each dot with different shape and volume represents the prediction error with respect to the parameter value it locates. When the prediction error goes larger, the color of the dot becomes from red to blue and its volume turns bigger. Different subfigures show different prediction error distribution over different types of the input pulses. It can be pretty intriguing that the prediction loss under low attenuation, large non-linearity and large dispersion whose second-order phase propagation constant is positive tends to be the largest as the shapes of dots in the front are larger.Other cases with large prediction error often when either parameter falls on the boundary like the edges of the parameter space.

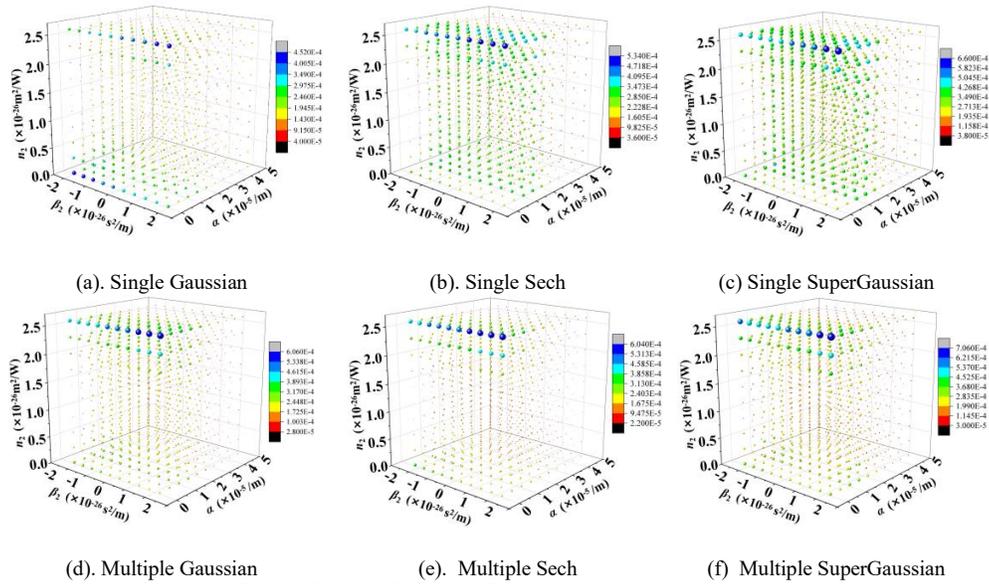

(a). Single Gaussian  (b). Single Sech  (c) Single SuperGaussian

(d). Multiple Gaussian  (e). Multiple Sech  (f) Multiple SuperGaussian

Fig. 11. Final prediction loss for the whole parameter space

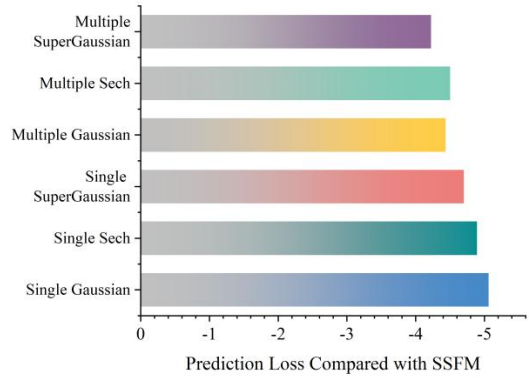

Fig. 10. Logarithmic final mean error loss of whole parameter space

In order to have a intuitive cognition of prediction accuracy of the model, we also conduct the computational consistency between our proposed model and the broadly adopted SSFM. The consistency is measured by calculated the average mean square error of each point of the waveform between our model and SSFM over the whole parameter space whose results can be vividly seen in Fig. 11. The total maximum average MSE does not exceed $.10^{-4}$ which is relatively high. The error different between different measurements may due to the intrinsic prediction error of SSFM.

### 3.5 Computational Complexity Comparison

Two companions of computational complexity should be conducted as to not only highlight the model's prominent advantages in fast predicting, but also reveal the time distribution differences of the whole model training and testing over different initial conditions. For the comparison of training and testing time distributions on different initial conditions, direct computing time is selected as the standard since the model is trained on Tesla T4 TPU all for those six different initial conditions. For the comparison of different methods in computational complexity, the scale of multiplication and addition (MAC) is calculated since different model tend to be operated on different hardware platforms.

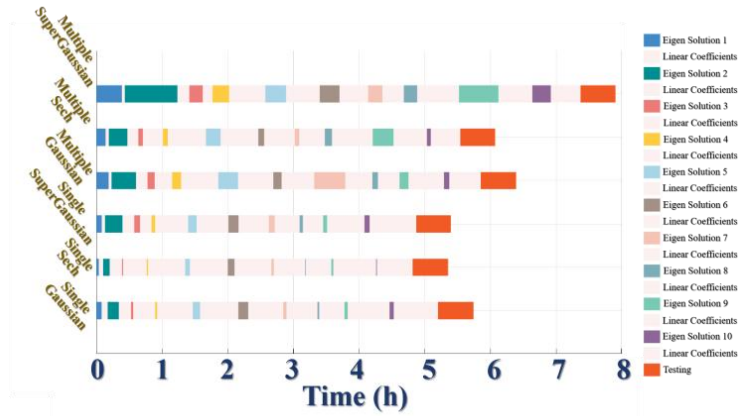

Fig. 12. Time consumption of the model

The computation complexity calculation and comparison of SSFM, the previously proposed principle driven fiber model and the parameterized fiber model proposed in this manuscript are conducted as well. Before comparison, several assumptions should be made. The task for calculate is a parameterized NLSE whose scale of parameter space equals $N$. Both coordinates of time and distance are discretized into $M_t$ and $M_\zeta$ respectively. The longest transmission distance is $L_{max}$. For comparison fairness, both principle driven fiber model and the inserted fiber model inside the parameterized model posses $K$ hidden layers, each layer contains $P$ neurons. For the parameterized fiber model, the maximum basis is determined to be $N_b$. The length of computational unit equal $L_u$. Under these assumptions, the computational complexity of SSFM can be calculated as

$$C_{SSFM} = O\left(\frac{TL_{max}}{L_u}\left[4M_t \log_2 M_t + N_{dispersion} + N_{nonlinear}\right]\right) \quad (10\text{-}A)$$

The complexity of the previously proposed fiber model should be

$$C_F = O\left(2TP + T(K-1)P^2\right) \quad (10\text{-}B)$$

And the complexity of the parameterized fiber model equals

$$C_{PF} = O\left(2N_b P + N_b(K-1)P^2 + 2N_b\right) \quad (10\text{-}C)$$

Since $N \gg N_b$, it can be easily concluded that the computational complexity of the parameterized fiber model is only the $N_b/N$ of that of the previously proposed fiber model.

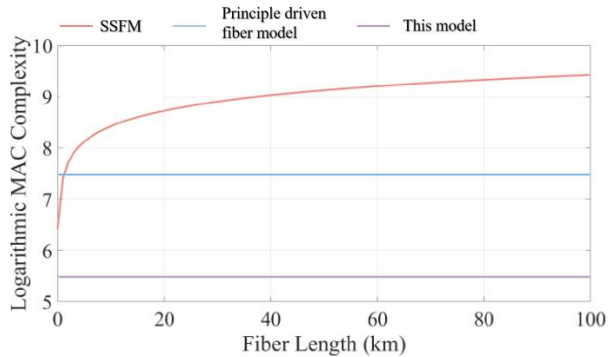

Fig. 13. complexity comparisons between three methods

## 4. Conclusions and discussions

In this manuscript, the parameterized principle driven fiber model is proposed. By taken into the use of both previously proposed fiber model to obtain the basis and the reduced basis expansion method to linearly combine those basis as to approximate the universal solution, any solutions with respect to the parameter space can be expressed in a relatively high accuracy.

Both signal pulse input and multiple pulse input are utilized to demonstrate the fidelity of the model. As can be seen from the section of simulation results and analysis, the predictions reaches relatively good acceptable levels though the prediction errors are relatively higher for multiple pulses input due to the pulses changing characteristics according to the analysis.

Several prominent advantages can be obtained when taking the model into use. First, this model can keep the better balance between prediction accuracy and physical backgrounds since it takes NLSE to design its loss function. Second, compared with other data driven models, this model can still be effectively trained without the needs to collect large scale of transmitted signals before hand. Last but not least, this model can solve pulse transmission at different fiber conditions all without retraining the whole model. By numerically analysis, when completing the task containing 1000 parameters with 10 maximum basis, this model can save 90% computational complexity compared with the previously proposed fiber model, not to mention that of SSFM. Future research plan will to further generalize this model into multi-mode fibers etc.

**Funding.** the National Key R&D Program of China (2022YFB2903600); the Youth Fund of the National Natural Science Foundation (NSFC) of China under Grant 62301275; Key Laboratory of Radar Imaging and Mircowave Photonics (Nanjing University of Aeronautics and Astronautics), Ministry of Education under Grant NJ20230003; the Research Start-up Fund of Nanjing University of Posts and Telecommunications under Grant NY223032.

**Disclosures.** The authors declare no conflicts of interests.

**Data availability.** Data can be available under reasonable requests.